\documentclass[10pt,twocolumn,letterpaper]{article}
\usepackage{cvpr}
\usepackage{times}
\usepackage{epsfig}
\usepackage{graphicx}
\usepackage{amsmath}
\usepackage{amssymb}
\usepackage{algorithmic}
\usepackage{float}
\usepackage{subcaption}
\usepackage{color}
\usepackage{xcolor}

\usepackage{algorithm}
\usepackage[square,numbers,sort]{natbib}
\usepackage[font=footnotesize,labelfont=bf]{caption}
\floatstyle{plain}
\newfloat{myalgo}{tbhp}{mya}
{\begin{myalgo}[#1]
\centering
\begin{minipage}{8cm}
\begin{algorithm}[H]}%
{\end{algorithm}
\end{minipage}
\end{myalgo}}

\def\sharedaffiliation{%
\end{tabular}
\begin{tabular}{c}}

\def\1n{\mathbf{1}_n}
\def\0{\mathbf{0}}
\def\1{\mathbf{1}}

\newcommand{\minisection}[1]{\noindent \textbf{#1}}
\def\posemachine{pose machine}

\usepackage[pagebackref=true,breaklinks=true,letterpaper=true,colorlinks,bookmarks=false]{hyperref}

\cvprfinalcopy 

\def\cvprPaperID{1074} 

\ifcvprfinal\fi

\begin{document}
\title{Convolutional Pose Machines}


\author{
Shih-En Wei \\ \texttt{\small shihenw@cmu.edu}
\and
Varun Ramakrishna \\ \texttt{\small vramakri@cs.cmu.edu}
\and
Takeo Kanade \\ \texttt{\small Takeo.Kanade@cs.cmu.edu}
\and
Yaser Sheikh \\ \texttt{\small yaser@cs.cmu.edu}
\sharedaffiliation
 The Robotics Institute \\
 Carnegie Mellon University
}


\maketitle
\begin{abstract}

Pose Machines provide a 
sequential prediction framework 
for 
learning 
rich implicit spatial models.
%
In this work we show a systematic design for how convolutional networks can be incorporated into the pose machine framework for learning image features and image-dependent spatial models for the task of pose estimation.
The contribution of this paper is to implicitly model long-range dependencies between variables in structured prediction tasks such as articulated pose estimation. We achieve this by designing a sequential architecture composed of convolutional networks that directly operate on belief maps from previous stages, producing increasingly refined estimates for part locations, without the need for explicit graphical model-style inference.
Our approach addresses the characteristic difficulty of vanishing gradients during training by providing a natural learning objective function that enforces intermediate supervision, thereby replenishing back-propagated gradients and conditioning the learning procedure. We demonstrate state-of-the-art performance and outperform competing methods on standard benchmarks including the MPII, LSP, and FLIC datasets.

%
\end{abstract}


\section{Introduction}


We introduce \emph{Convolutional Pose Machines (CPMs)} for the task of articulated pose estimation. CPMs inherit the benefits of the \emph{\posemachine} \cite{Ramakrishna2014posemachines} architecture---the implicit learning of long-range dependencies between image and multi-part cues, tight integration between learning and inference, a modular sequential design---and combine them with the advantages afforded by convolutional architectures: the ability to learn feature representations for both image and spatial context directly from data; a differentiable architecture that allows for globally joint training with backpropagation; and the ability to efficiently handle large training datasets.

CPMs consist of a sequence of convolutional networks that repeatedly produce 2D belief maps~\footnote{We use the term \emph{belief} in a slightly loose sense, however the belief maps described are closely related to beliefs produced in message passing inference in graphical models. The overall architecture can be viewed as an unrolled mean-field message passing inference algorithm~\cite{ross2011} that is learned end-to-end using backpropagation.} 
for the location of each part. At each stage in a CPM, image features and the belief maps produced by the previous stage are used as input. The belief maps provide the subsequent stage an expressive non-parametric encoding of the spatial uncertainty of location for each part, allowing the CPM to learn rich image-dependent spatial models of the relationships between parts. Instead of explicitly parsing such belief maps either using graphical models~\cite{tompson2014joint,tompson2015cvpr,pishchulin2015deepcut} or specialized post-processing steps~\cite{toshev2013deeppose,tompson2015cvpr}, we learn convolutional networks that directly operate on intermediate belief maps and learn implicit image-dependent spatial models of the relationships between parts. The overall proposed multi-stage architecture is fully differentiable and therefore can be trained in an end-to-end fashion using backpropagation.

\begin{figure}[t!]
    \centering
    \includegraphics[width=1\columnwidth]{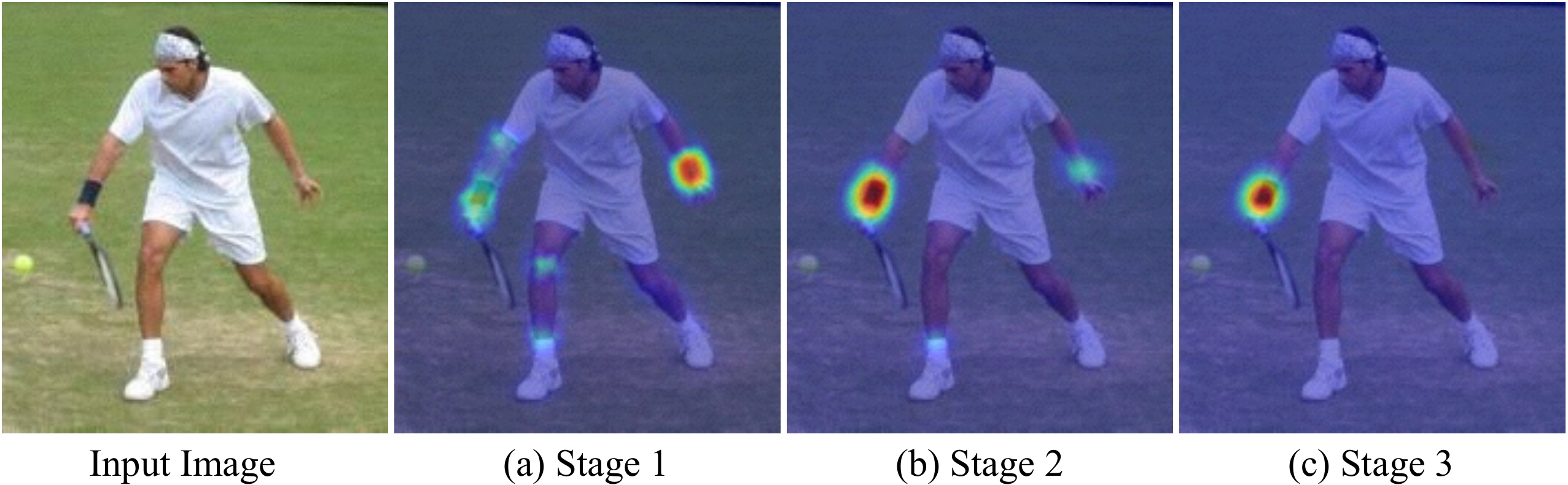}
    \caption{A \textbf{Convolutional Pose Machine} consists of a sequence of predictors trained to make dense predictions at each image location. Here we show the increasingly refined estimates for the location of the \emph{right elbow} in each stage of the sequence.
(a) Predicting from local evidence often causes confusion. (b) Multi-part context helps resolve ambiguity. (c) Additional iterations help converge to a certain solution.}
    \label{fig:teaser}
\end{figure}

At a particular stage in the CPM, the spatial context of part beliefs provide strong disambiguating cues to a subsequent stage. As a result, each stage of a CPM produces belief maps with increasingly refined estimates for the locations of each part (see Figure~\ref{fig:teaser}). 
In order to capture long-range interactions between parts, the design of the network in each stage of our sequential prediction framework is motivated by the goal of achieving a large receptive field on both the image and the belief maps. We find, through experiments, that large receptive fields on the belief maps are crucial for learning long range spatial relationships and result in improved accuracy.

Composing multiple convolutional networks in a CPM results in an overall network with many layers that is at risk of the problem of
\emph{vanishing gradients}~\cite{hochreiter2001gradient,bradley2010learning,glorot2010understanding,bengio1994learning} during learning. 
This problem can occur because back-propagated gradients diminish in strength as they are propagated through the many layers of the network. 
While there exists recent work~\footnote{New results have shown that using skip connections with identity mappings~\cite{he15arxiv} in so-called residual units also aids in addressing vanishing gradients in ``very deep'' networks. We view this method as complementary and it can be noted that our modular architecture easily allows us to replace each stage with the appropriate residual network equivalent.} which shows that supervising very deep networks at intermediate layers aids in learning~\cite{lee2014deeply,szegedy2014going}, they have mostly been restricted to classification problems.
In this work, we show how for a structured prediction problem such as pose estimation, CPMs naturally suggest a systematic framework that replenishes gradients and guides the network to produce increasingly accurate belief maps  by enforcing intermediate supervision periodically through the network. We also discuss different training schemes of such a sequential prediction architecture. 
Our main contributions are (a) learning implicit spatial models via a sequential composition of convolutional architectures and (b) a systematic approach to designing and training such an architecture  to learn both image features and image-dependent spatial models for structured prediction tasks, without the need for any graphical model style inference. We achieve state-of-the-art results on standard benchmarks including the MPII, LSP, and FLIC datasets, and analyze the effects of jointly training a multi-staged architecture with repeated intermediate supervision.

\section{Related Work}
The classical approach to articulated pose estimation is the \textbf{pictorial structures} model \cite{fh2005pictorial,ramanan2005strike,Andriluka2010,Andriluka2009,pishchulin13iccv,pishchulin2013poselet,yang2011articulated,johnson2010clustered} in which spatial correlations between parts of the body are expressed as a tree-structured graphical model with kinematic priors that couple connected limbs. These methods have been successful on images where all the limbs of the person are visible, but are prone to characteristic errors such as double-counting image evidence, which occur because of correlations between variables that are not captured by a tree-structured model. The work of Kiefel et al. \cite{kiefel2014human} is based on the pictorial structures model but differs in the underlying graph representation. \textbf{Hierarchical models} \cite{tian2012exploring,sun2011articulated} represent the relationships between parts at different scales and sizes in a hierarchical tree structure. The underlying assumption of these models is that larger parts (that correspond to full limbs instead of joints) can often have discriminative image structure that can be easier to detect and consequently help reason about the location of smaller, harder-to-detect parts. 
\textbf{Non-tree models} \cite{wang2008multiple,sigal2006measure,lan2005beyond,karlinsky2012using,Dantone2013} incorporate interactions that introduce loops to augment the tree structure with additional edges that capture symmetry, occlusion and long-range relationships. These methods usually have to rely on approximate inference during both learning and at test time, and therefore have to trade off accurate modeling of spatial relationships with models that allow efficient inference, often with a simple parametric form to allow for fast inference. In contrast, methods based on a \textbf{sequential prediction} framework \cite{Ramakrishna2014posemachines} learn an \emph{implicit} spatial model with potentially complex interactions between variables by directly training an inference procedure, as in \cite{munoz2010,ross2011,tu2010PAMI,Pinheiro14recurrentconvolutional}. 

There has been a recent surge of interest in models that employ \textbf{convolutional architectures} for the task of articulated pose estimation \cite{ouyang2014multi,tompson2015cvpr,tompson2014joint,Chen_NIPS14,
carreia2015human,pishchulin2015deepcut,pfister2015flowing}. 
Toshev et al. \cite{toshev2013deeppose} take the approach of directly regressing the Cartesian coordinates using a standard convolutional architecture \cite{krizhevsky2012imagenet}.
%
%
Recent work regresses image to confidence maps, and resort to graphical models, which require hand-designed energy functions or heuristic initialization of spatial probability priors, to remove outliers on the regressed confidence maps. Some of them also utilize a dedicated network module for precision refinement \cite{tompson2015cvpr,pishchulin2015deepcut}.
In this work, we show the regressed confidence maps are suitable to be inputted to further convolutional networks with large receptive fields to learn implicit spatial dependencies without the use of hand designed priors, and achieve state-of-the-art performance over all precision region without careful initialization and dedicated precision refinement. Pfister et al. \cite{pfister2015flowing} also used a network module with large receptive field to capture implicit spatial models.
Due to the differentiable nature of convolutions, our model can be globally trained, where Tompson et al. \cite{tompson2014joint} and Steward et al. \cite{steward2015endtoend} also discussed the benefit of joint training.

%
%

Carreira et al.~\cite{carreia2015human} train a deep network that iteratively improves part detections using error feedback but use a cartesian representation as in~\cite{toshev2013deeppose} which does not preserve spatial uncertainty and results in lower accuracy in the high-precision regime.
%
In this work, we show how the sequential prediction framework takes advantage of the preserved uncertainty in the confidence maps to encode the rich spatial context, with enforcing the intermediate local supervisions to address the problem of vanishing gradients.

\section{Method}

\begin{figure*}[t!]
    \centering
    \includegraphics[width=1\textwidth]{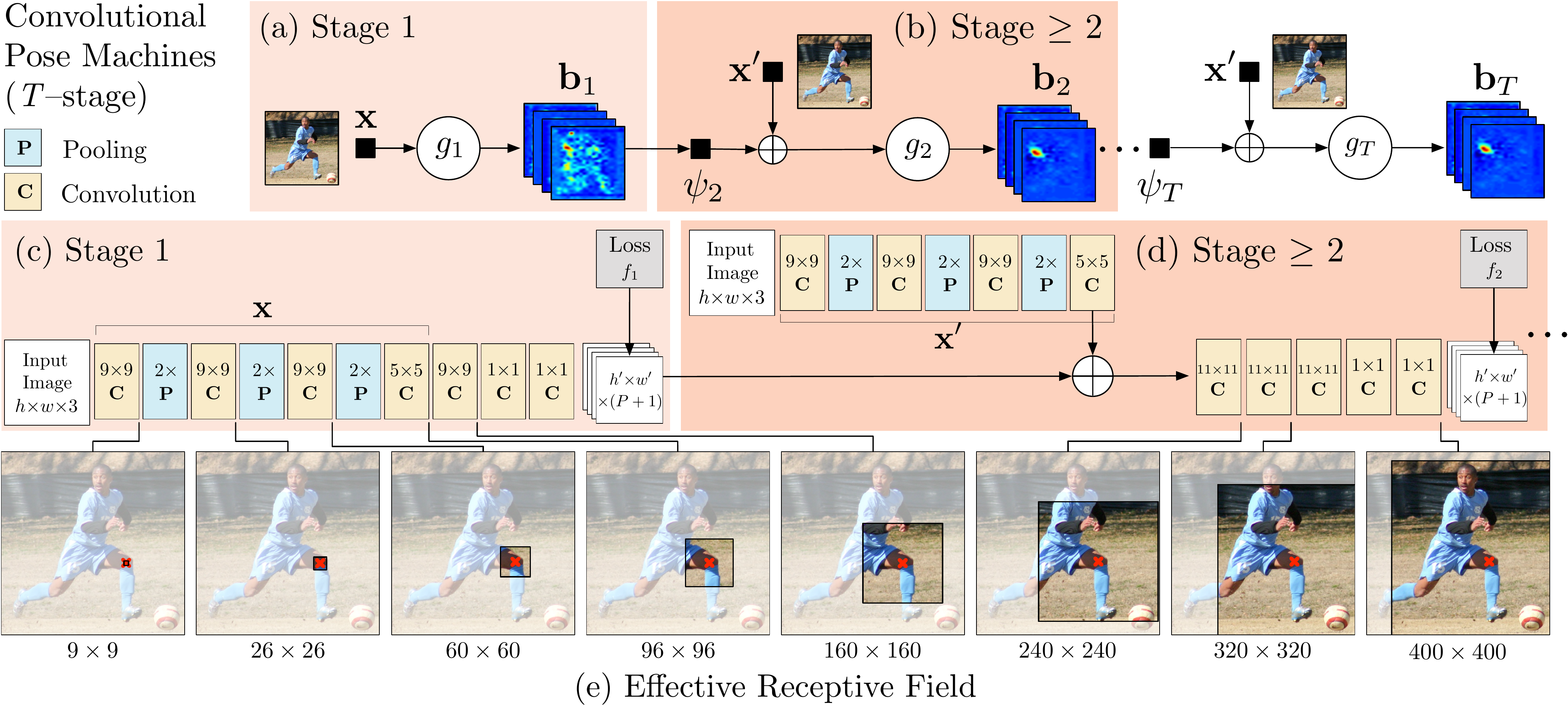}
    \caption{\textbf{Architecture and receptive fields of CPMs.} We show a convolutional architecture and receptive fields across layers for a CPM with any $T$ stages. The \posemachine~\cite{Ramakrishna2014posemachines} is shown in insets (a) and (b), and the corresponding convolutional networks are shown in insets (c) and (d). Insets (a) and (c) show the architecture that operates only on image evidence in the first stage. Insets (b) and (d) shows the architecture for subsequent stages, which operate both on image evidence as well as belief maps from preceding stages. The architectures in (b) and (d) are repeated for all subsequent stages (2 to $T$). The network is locally supervised after each stage using an intermediate loss layer that prevents vanishing gradients during training. Below in inset (e) we show the effective receptive field on an image (centered at left knee) of the architecture, where the large receptive field enables the model to capture long-range spatial dependencies such as those between head and knees. (Best viewed in color.)}
    \label{fig:arch_half}
\end{figure*} 

\subsection{Pose Machines}
We denote the pixel location of the $p$-th anatomical landmark (which we refer to as a part),
$Y_p \in \mathcal{Z} \subset \mathbb{R}^2$, where $\mathcal{Z}$ is the set of all $(u, v)$ locations in an image. 
Our goal is to predict the image locations 
$Y=(Y_1,\ldots,Y_{P})$ 
for all $P$ parts. 
A \posemachine~\cite{Ramakrishna2014posemachines} (see Figure \ref{fig:arch_half}a and \ref{fig:arch_half}b) consists of a sequence of multi-class predictors, 
$g_t(\cdot)$,
that are trained to predict the location of each part in each level of the hierarchy. In each \emph{stage} $t \in \{ 1 \ldots T\}$, the classifiers 
$g_t$
predict beliefs for assigning a location to each part $Y_p = z, ~\forall z \in \mathcal{Z},$ based on features extracted from the image at the location $z$ denoted by $\mathbf{x}_z \in \mathbb{R}^d$ and contextual information from the preceding classifier in the neighborhood around each $Y_p$ in stage $t$. 
A classifier 
in the first stage $t = 1$, therefore produces the following belief values:
\begin{equation}
g_{1}(\mathbf{x}_z) \rightarrow \left\{ b_1^p( Y_p = z)\right\}_{p \in \{0 \ldots P\}},
\end{equation}

\noindent where 
$b^{p}_1(Y_p = z)$
is the score predicted by the classifier 
$g_1$
for assigning the $p^{\mathrm{th}}$ part 
in the first stage at image location $z$. We represent all the beliefs of part $p$ 
evaluated at every location $z=(u,v)^T$ in the image as 
$\mathbf{b}^p_t \in \mathbb{R}^{w \times h}$, 
where $w$ and $h$ are the width and height of the image, respectively. That is,
\begin{equation}
\mathbf{b}^p_t[u,v] = b^{p}_t(Y_p = z).
\end{equation}
For convenience, we denote the collection of belief maps for all the parts 
as 
$\mathbf{b}_t \in \mathbb{R}^{w \times h \times (P+1)}$ ($P$ parts plus one for background).

In subsequent stages, the classifier predicts a belief for assigning a location to each part $Y_p = z, ~\forall z \in \mathcal{Z},$ based on (1) features of the image data $\mathbf{x}^t_z \in \mathbb{R}^d$ again, and (2) contextual information from the preceeding classifier in the neighborhood around each $Y_p$:
\begin{equation}
    g_t \left(\mathbf{x}'_z, \psi_t(z, \mathbf{b}_{t-1}) \right) \rightarrow \left\{ b_t^p( Y_p = z)\right\}_{p \in \{0 \ldots P+1\}},
    \label{eqn:predoutput}
\end{equation}

\noindent where 
$\psi_{t>1}(\cdot)$ is a mapping from the beliefs $\mathbf{b}_{t-1}$ to context features. In each stage, the computed beliefs provide an increasingly refined estimate for the location of each part. Note that we allow image features $\mathbf{x}'_z$ for subsequent stage to be different from the image feature used in the first stage $\mathbf{x}$.
The \posemachine~proposed in~\cite{Ramakrishna2014posemachines} used boosted random forests for prediction 
($\{g_t\}$),
fixed hand-crafted image features across all stages ($\mathbf{x}' = \mathbf{x}$), and fixed hand-crafted context feature maps ($ \psi_t(\cdot)$) to capture spatial context across all stages.

\subsection{Convolutional Pose Machines}
We show how the prediction and image feature computation modules of a pose machine can be replaced by a deep convolutional architecture allowing for both image and contextual feature representations to be learned directly from data. Convolutional architectures also have the advantage of being completely differentiable, thereby enabling end-to-end joint training of all stages of a CPM. 
We 
describe our design for a CPM that combines the advantages of deep convolutional architectures with the implicit spatial modeling afforded by the pose machine framework.

 \subsubsection{Keypoint Localization Using Local Image \\ Evidence}
The first stage of a convolutional pose machine predicts part beliefs from only local image evidence. Figure \ref{fig:arch_half}c shows the network structure used for part detection from local image evidence using a deep convolutional network. The evidence is \emph{local} because the receptive field of the first stage of the network is constrained to a 
small patch around the output pixel location.
We use a network structure composed of five convolutional layers followed by two $1 \times 1$ convolutional layers which results in a fully convolutional architecture \cite{long_shelhamer_fcn}.
In practice, to achieve certain precision, we normalize input cropped images to size $368\times 368$ (see Section~\ref{subset:Quantitative} for details), and the receptive field of the network shown above is $160 \times 160$ pixels. The network can effectively be viewed as sliding a deep network across an image and regressing from the local image evidence in each $160\times 160$ image patch to a $P + 1$ sized output vector that represents a score for each part at that image location.


\begin{figure}[t!]
\centering
\includegraphics[width=\linewidth]{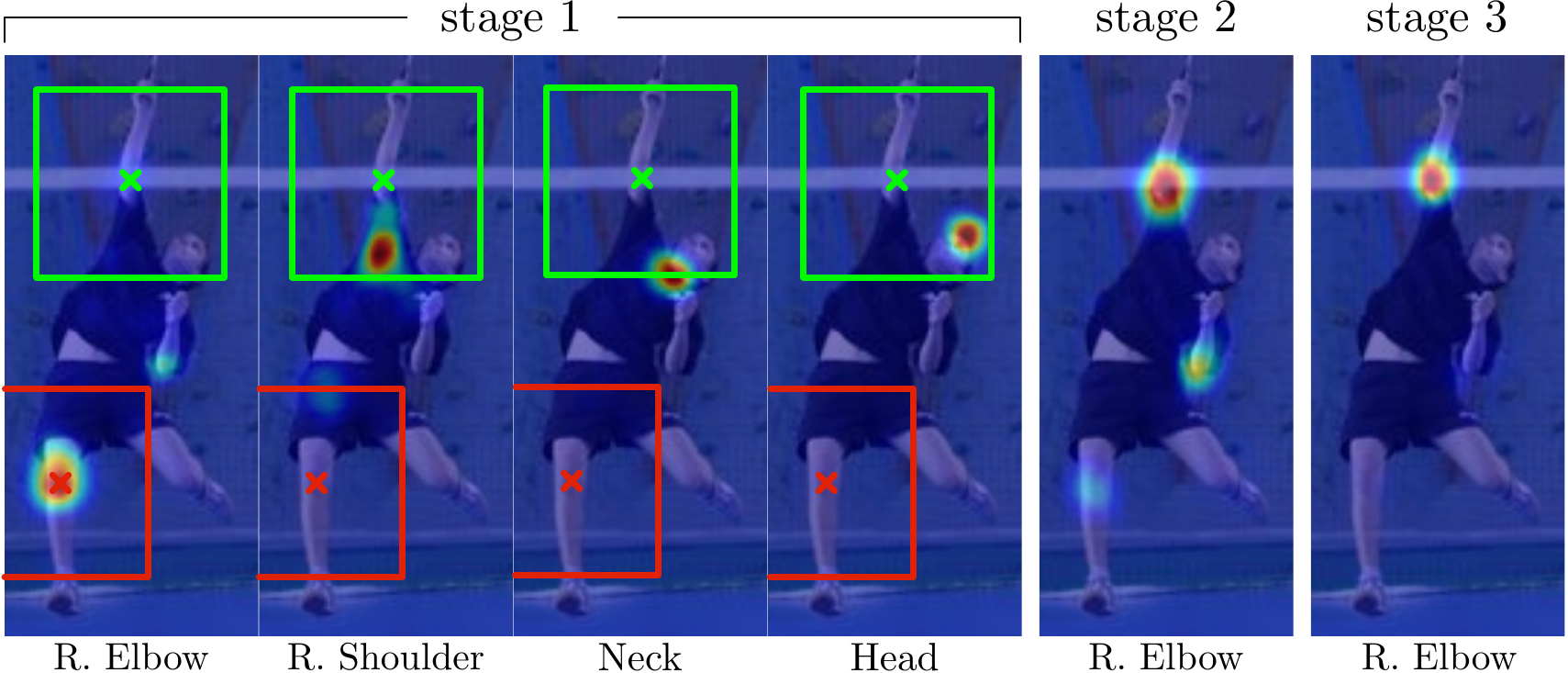}
\caption{\textbf{Spatial context from belief maps} of easier-to-detect parts can provide strong cues for localizing difficult-to-detect parts. The spatial contexts from shoulder, neck and head can help eliminate wrong (red) and strengthen correct (green) estimations on the belief map of \emph{right elbow} in the subsequent stages.}
\label{fig:contextfeaturesgood}
\end{figure}

\subsubsection{Sequential Prediction with Learned Spatial\\ Context Features}
While the detection rate on landmarks with consistent appearance, such as the head and shoulders, can be favorable, the accuracies are often much lower for landmarks lower down the kinematic chain of the human skeleton due to their large variance in configuration and appearance. The landscape of the belief maps around a part location, albeit noisy, can, however, be very informative. 
Illustrated in Figure \ref{fig:contextfeaturesgood}, when detecting challenging parts such as right elbow, the belief map for right shoulder with a sharp peak can be used as a strong cue. 
%
A predictor in subsequent stages ($g_{t>1}$) can use the spatial context ($\psi_{t>1}(\cdot)$) of the noisy belief maps in a region around the image location $z$ and improve its predictions by leveraging the fact that parts occur in consistent geometric configurations.  In the second stage of a \posemachine, the classifier 
$g_2$
accepts as input the image features 
$\mathbf{x}^2_{z}$
and features computed on the beliefs via the feature function $\psi$ for each of the parts in the previous stage. The feature function $\psi$ serves to encode the landscape of the belief maps from the previous stage in a spatial region around the location $z$ of the different parts.
For a convolutional pose machine, we do not have an explicit function that computes context features. Instead, we define $\psi$ as being the receptive field of the predictor on the beliefs from the previous stage.

The design of the network is guided by achieving a receptive field at the output layer of the second stage network that is large enough to allow the learning of potentially complex and long-range correlations between parts. By simply supplying features on the outputs of the previous stage (as opposed to specifying potential functions in a graphical model), the convolutional layers in the subsequent stage allow the classifier to freely combine contextual information by picking the most predictive features. 
The belief maps from the first stage are generated from a network that examined the image locally with a small receptive field.
\begin{figure}[t!]
    \centering
    \includegraphics[width=0.49\columnwidth]{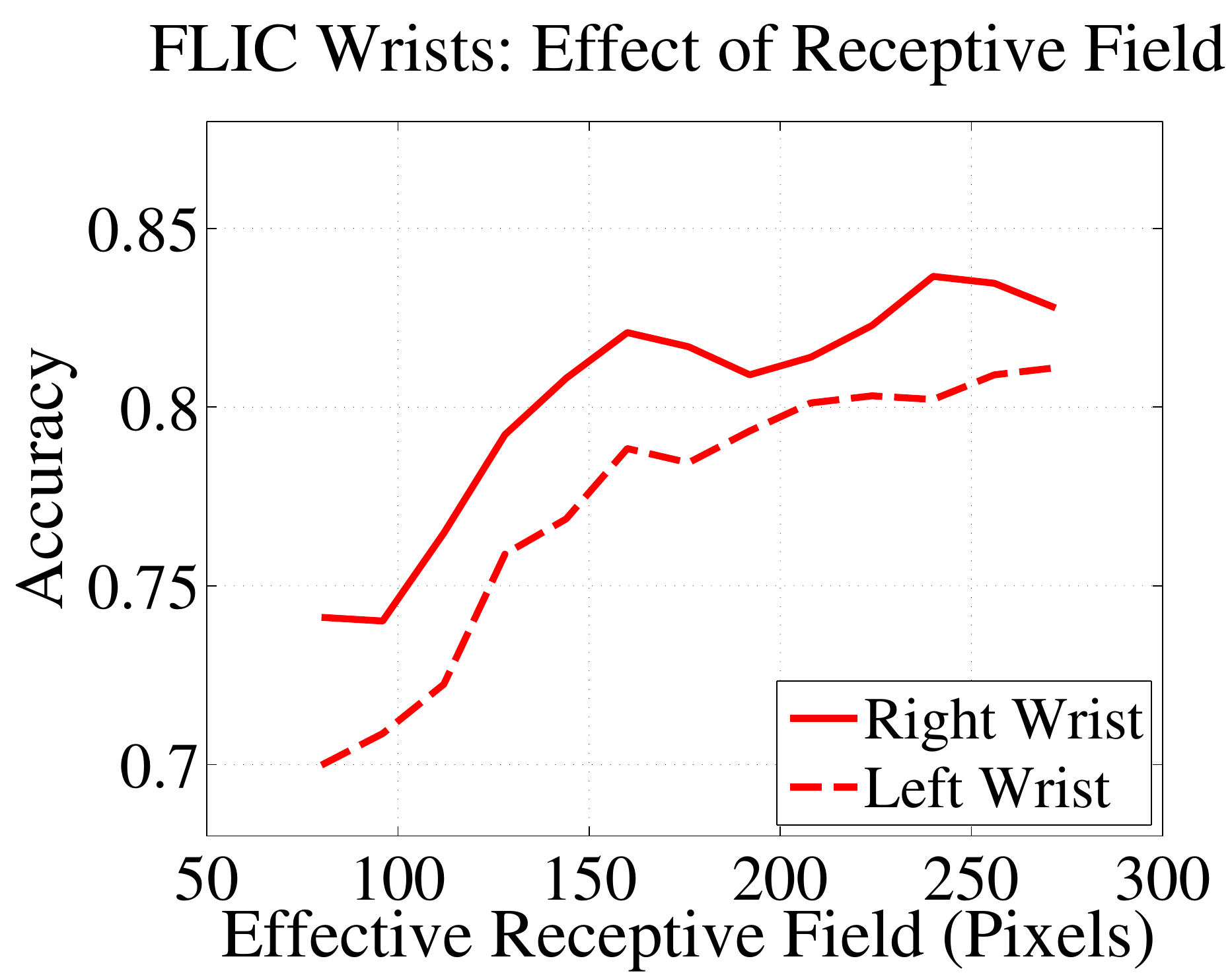}
    \includegraphics[width=0.49\columnwidth]{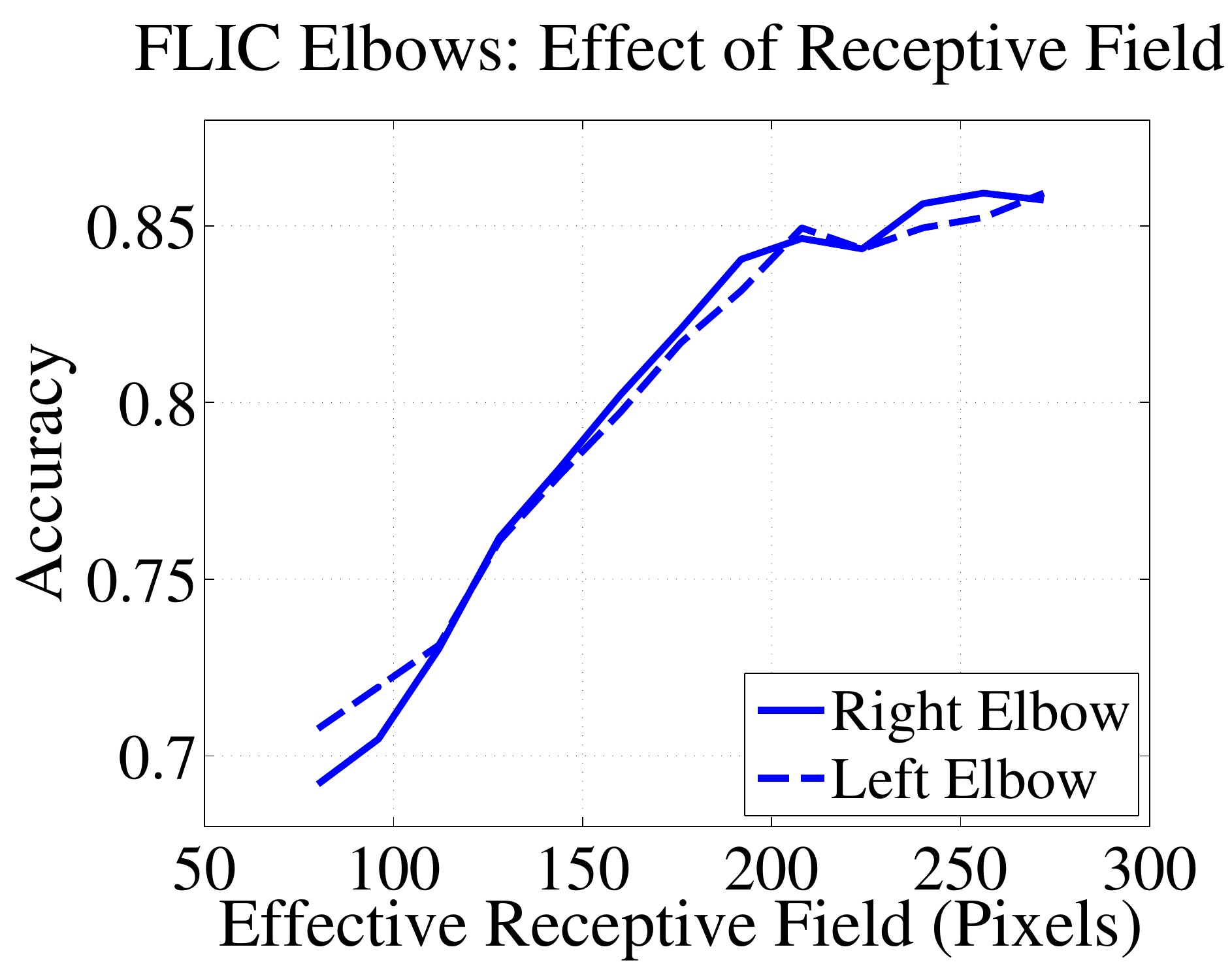}
    \caption{\textbf{Large receptive fields for spatial context.} We show that networks with large receptive fields are effective at modeling long-range spatial interactions between parts. Note that these experiments are operated with smaller normalized images than our best setting.}
    \label{fig:flic_kernels}
\end{figure}
In the second stage, we design a network that drastically increases the equivalent receptive field. 
Large receptive fields can be achieved either by pooling at the expense of precision, increasing the kernel size of the convolutional filters at the expense of increasing the number of parameters, or by increasing the number of convolutional layers at the risk of encountering vanishing gradients during training. 
Our network design and corresponding receptive field for the subsequent stages ($t \geq 2$) is shown in Figure \ref{fig:arch_half}d.
We choose to use multiple convolutional layers to achieve large receptive field on the $8\times$ downscaled heatmaps, as it allows us to be parsimonious with respect to the number of parameters of the model. We found that our stride-$8$ network performs as well as a stride-$4$ one even at high precision region, while it makes us easier to achieve larger receptive fields.
We also repeat similar structure for image feature maps to make the spatial context be image-dependent and allow error correction, following the structure of \posemachine.

\begin{figure*}[ht!]
    \centering
    \includegraphics[width=1\linewidth]{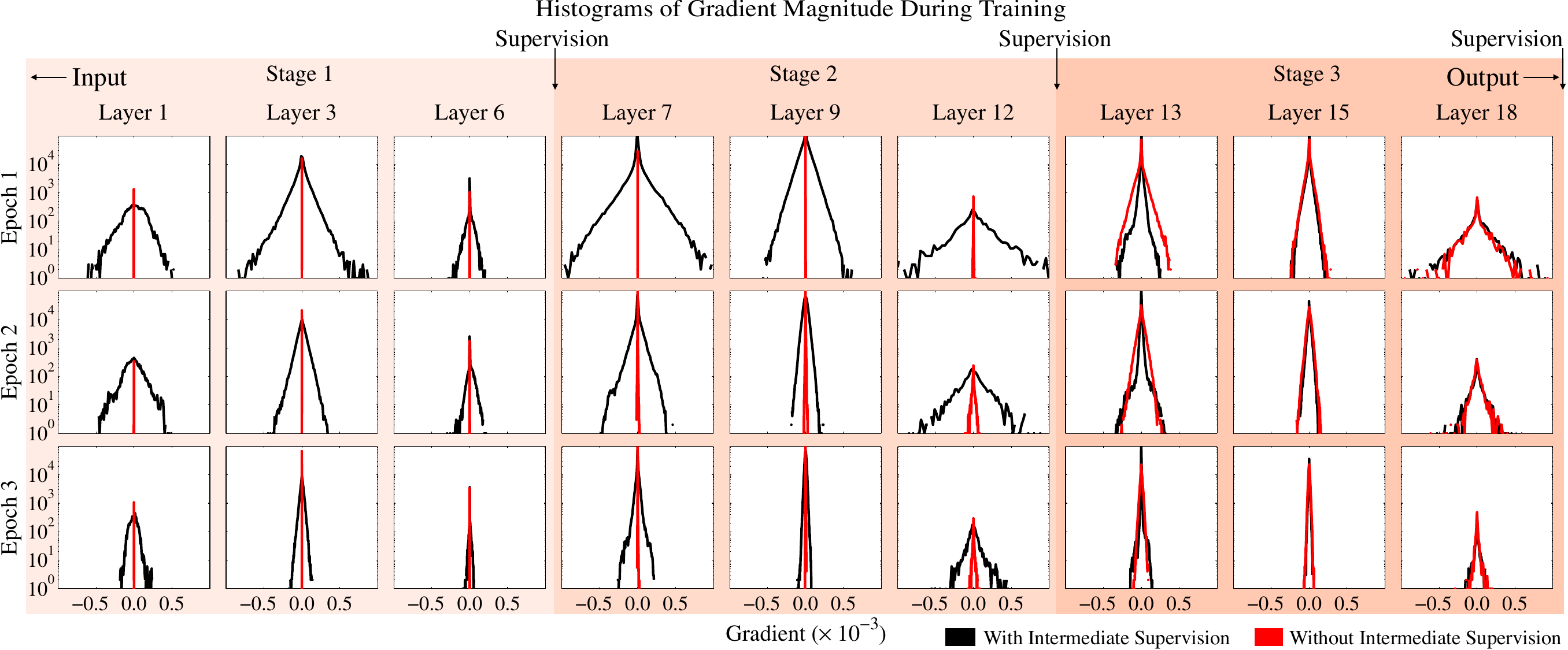}
    \caption{\textbf{Intermediate supervision addresses vanishing gradients.} We track the change in magnitude of gradients in layers at different depths in the network, across training epochs, for models with and without intermediate supervision. We observe that for layers closer to the output, the distribution has a large variance for both with and without intermediate supervision; however as we move from the output layer towards the input, the gradient magnitude distribution peaks tightly around zero with low variance (the gradients \emph{vanish}) for the model without intermediate supervision. For the model with intermediate supervision the distribution has a moderately large variance throughout the network. At later training epochs, the variances decrease for all layers for the model with intermediate supervision and remain tightly peaked around zero for the model without intermediate supervision. (Best viewed in color)}
    \label{fig:gradient_change}
\end{figure*}

We find that accuracy improves with the size of the receptive field. In Figure \ref{fig:flic_kernels} we show the improvement in accuracy on the FLIC dataset \cite{sappmodec} as the size of the receptive field on the original image is varied by varying the architecture without significantly changing the number of parameters,  through a series of experimental trials on input images normalized to a size of $304 \times 304$. We see that the accuracy improves as the effective receptive field increases, and starts to saturate around $250$ pixels, which also happens to be roughly the size of the normalized object. This improvement in accuracy with receptive field size suggests that the network does indeed encode long range interactions between parts and that doing so is beneficial.
In our best performing setting in Figure~\ref{fig:arch_half}, we normalize cropped images into a larger size of $368 \times 368$ pixels for better precision, and the receptive field of the second stage output on the belief maps of the first stage is set to $31\times 31$, which is equivalently $400\times 400$ pixels on the original image, where the radius can usually cover any pair of the parts. With more stages, the effective receptive field is even larger. In the following section we show our results from up to $6$ stages.



\begin{figure*}[ht!]
    \begin{subfigure}{0.33\textwidth}
   	     \centering
         \includegraphics[width=\textwidth]{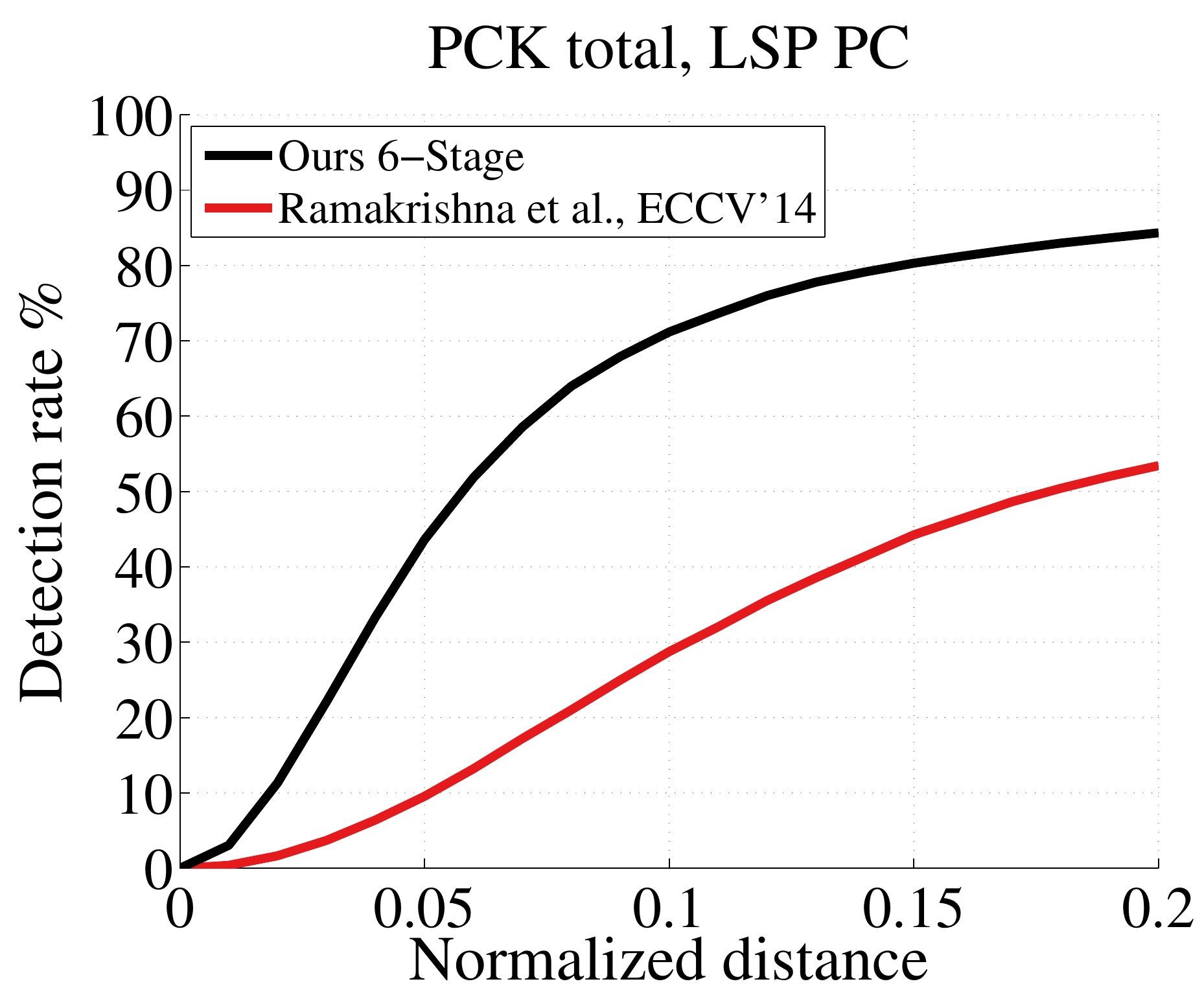}
         \caption{}
         \label{fig:comp_with_pm}
    \end{subfigure}
    \begin{subfigure}{0.33\textwidth}
        \centering
        \includegraphics[width=\textwidth]{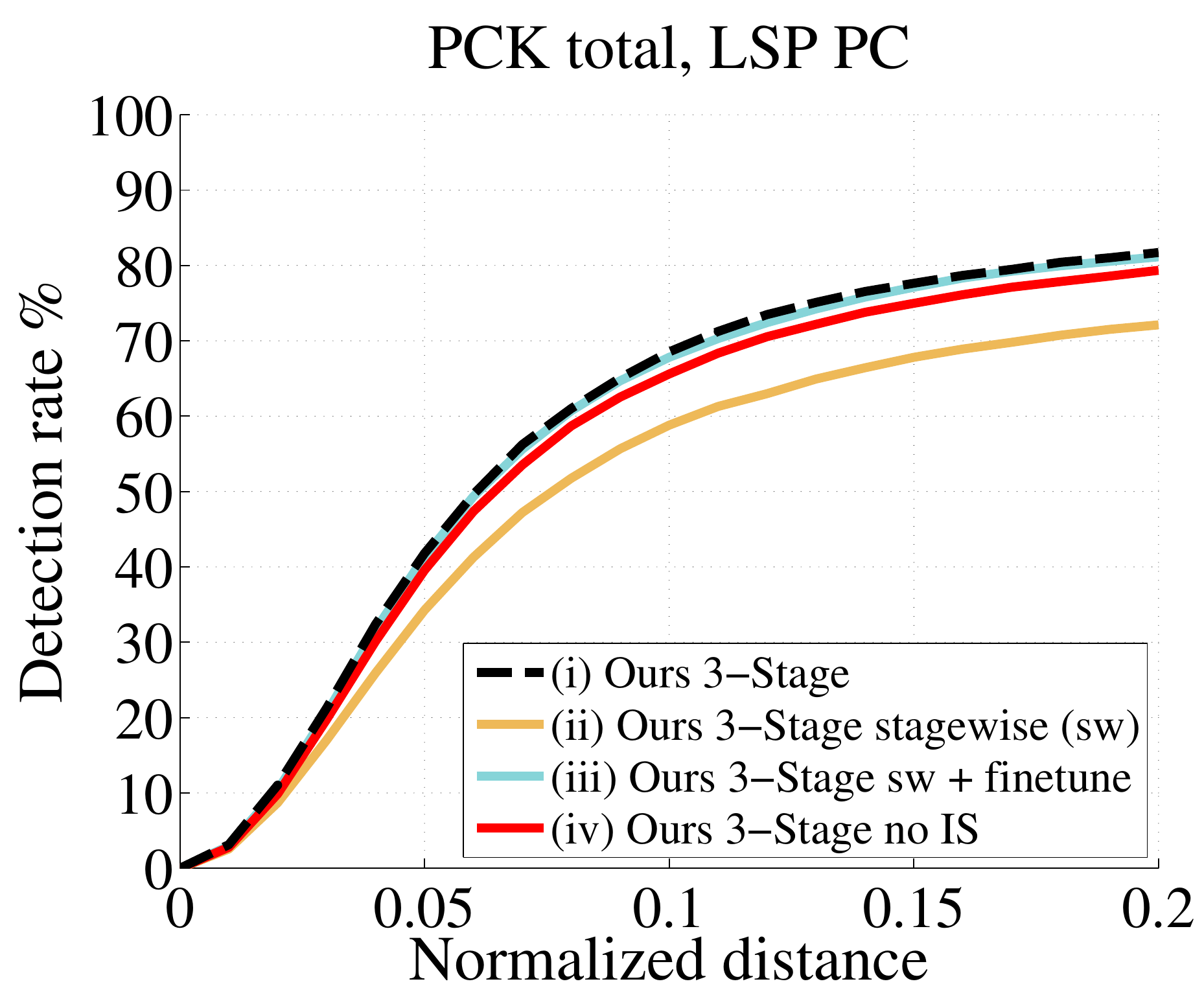}
        \caption{}
        \label{fig:compare_training}
    \end{subfigure}
    \begin{subfigure}{0.33\textwidth}
        \centering
        \includegraphics[width=\textwidth]{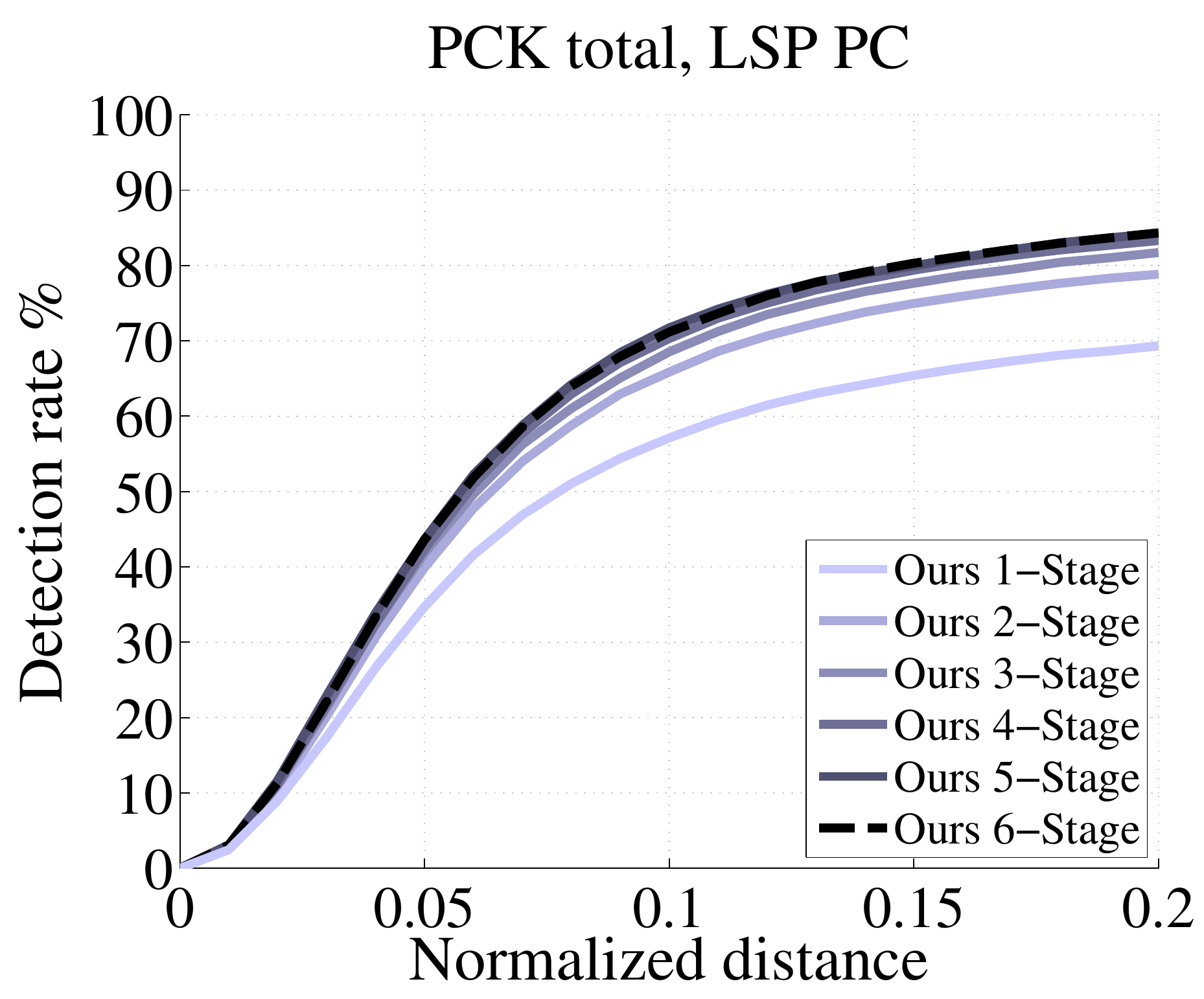}
        \caption{}
        \label{fig:compare_stages}
    \end{subfigure}
    \caption{\textbf{Comparisons on 3-stage architectures on the LSP dataset (PC)}: (a) Improvements over Pose Machine. (b) Comparisons between the different training methods. (c) Comparisons across each number of stages using joint training from scratch with intermediate supervision.} 
\end{figure*}

\subsection{Learning in Convolutional Pose Machines}
The design described above for a \posemachine~results in a deep architecture that can have a large number of layers.  Training such a network with many layers can be prone to the problem of \emph{vanishing gradients} \cite{bradley2010learning,glorot2010understanding,bengio1994learning} where, as observed 
by Bradley \cite{bradley2010learning} and Bengio et al. \cite{glorot2010understanding}, the magnitude of back-propagated gradients decreases in strength with the number of intermediate layers between the output layer and the input layer.


Fortunately, the sequential prediction framework of the \posemachine~provides a natural approach to training our deep architecture that addresses this problem. Each stage of the \posemachine~is trained to repeatedly produce the belief maps for the locations of each of the parts. We encourage the network to repeatedly arrive at such a representation by defining a loss function at the output of each stage $t$ 
that minimizes the $l_2$ distance between the predicted and ideal belief maps for each part. The ideal belief map for a part $p$ is written as 
$b^{p}_{*}(Y_p = z)$,
which are created by putting Gaussian peaks at ground truth locations of each body part $p$.
The cost function we aim to minimize at the output of each stage at each level is therefore given by:
 \begin{equation}
    f_t = \sum_{p = 1}^{P+1} \sum_{z\in \mathcal{Z}}  \| b^{p}_{t}(z) -  b^{p}_{*}(z)\|^{2}_{2}.
    \label{eqn:localobjective}
 \end{equation}

The overall objective for the full architecture is obtained by adding the losses at each stage and is given by:
\begin{equation}
    \mathcal{F} = \sum_{t = 1}^{T} f_t .
\label{eqn:fullobjective}
\end{equation}
We use standard stochastic gradient descend to jointly train all the $T$ stages in the network. To share the image feature $\mathbf{x}'$ across all subsequent stages, we share the weights of corresponding convolutional layers (see Figure~\ref{fig:arch_half}) across stages $t \geq 2$.


\section{Evaluation}

\subsection{Analysis}
\minisection{Addressing vanishing gradients.}
The objective in Equation \ref{eqn:fullobjective} describes a decomposable loss function that operates on different parts of the network (see Figure \ref{fig:arch_half}). Specifically, each term in the summation is applied to the network after each stage $t$ effectively enforcing supervision in intermediate stages through the network. Intermediate supervision has the advantage that, even though the full architecture can have many layers, it does not fall prey to the \emph{vanishing gradient} problem as the intermediate loss functions replenish the gradients at each stage.

We verify this claim by observing histograms of gradient magnitude (see Figure \ref{fig:gradient_change}) at different depths in the architecture across training epochs for models with and without intermediate supervision. In early epochs, as we move from the output layer to the input layer, we observe on the model \emph{without intermediate supervision}, the gradient distribution is tightly peaked around zero because of vanishing gradients. The model \emph{with intermediate supervision} has a much larger variance across all  layers, suggesting that learning is indeed occurring in all the layers thanks to intermediate supervision. We also notice that as training progresses, the variance in the gradient magnitude distributions decreases pointing to model convergence.

\minisection{Benefit of end-to-end learning.} 
We see in Figure \ref{fig:comp_with_pm} that replacing the modules of a pose machine with the appropriately designed convolutional architecture provides a large boost of $42.4$ percentage points over the previous approach of \cite{Ramakrishna2014posemachines} in the high precision regime (PCK@0.1) and  $30.9$ percentage points in the low precision regime (PCK@0.2).

\minisection{Comparison on training schemes.}
\label{sec:compare_learning} We compare different variants of training the network in Figure \ref{fig:compare_training} on the LSP dataset with person-centric (PC) annotations. To demonstrate the benefit of intermediate supervision with joint training across stages, we train the model in four ways:
(i) training from scratch using a global loss function that enforces intermediate supervision
(ii) stage-wise; where each stage is trained in a feed-forward fashion and stacked
(iii) as same as (i) but initialized with weights from (ii), and
(iv) as same as (i) but with no intermediate supervision. We find that network (i) outperforms all other training methods, showing that intermediate supervision and joint training across stage is indeed crucial in achieving good performance. The stagewise training in (ii) saturate at sub-optimal, and the jointly fine-tuning in (iii) improves from this sub-optimal to the accuracy level closed to (i), however with effectively longer training iterations.


\begin{figure*}[ht!]
    \centering
    \includegraphics[width=1\linewidth]{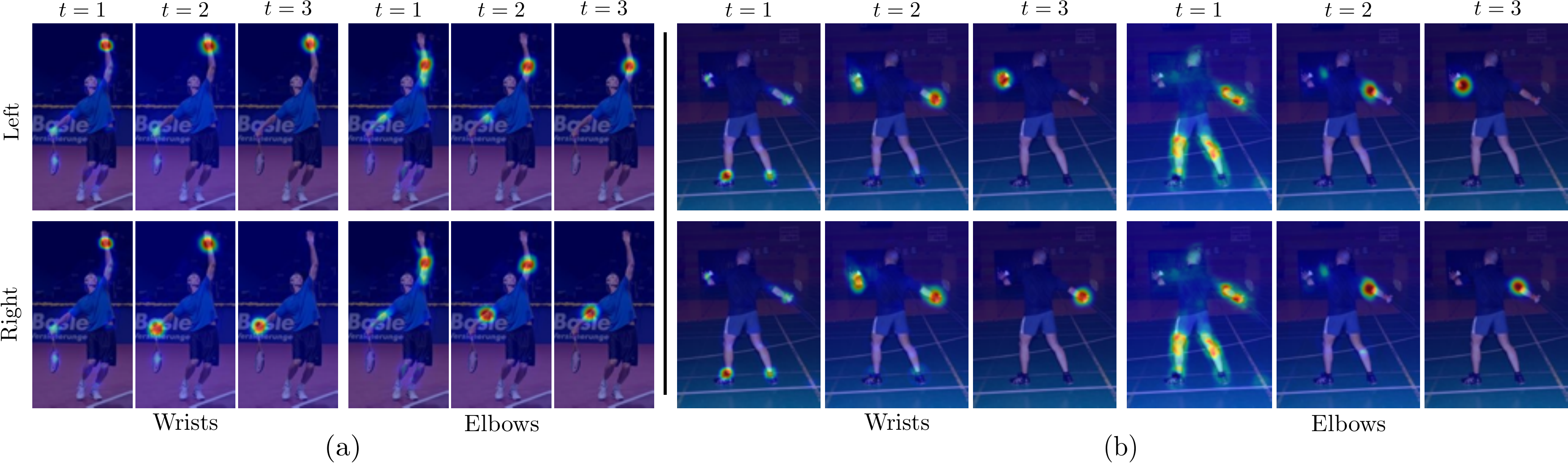}
    \vspace{-20pt}
    \caption{\textbf{Comparison of belief maps across stages} for the elbow and wrist joints on the LSP dataset for a 3-stage CPM.} 
    \label{fig:leeds_qual}
\end{figure*}

\begin{figure*}[ht!]
        \centering
        \includegraphics[width=0.965\textwidth]{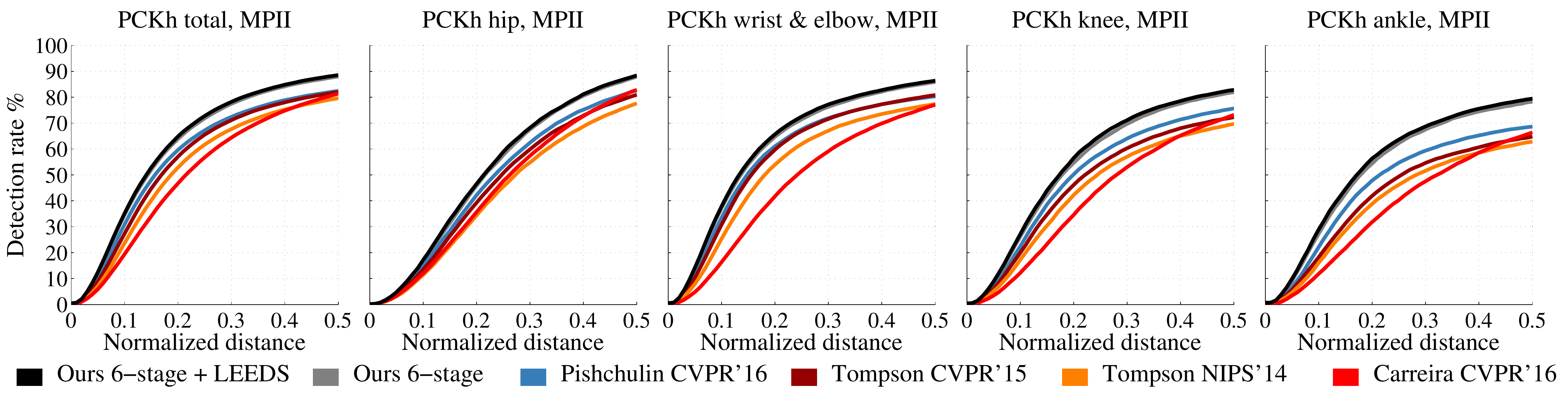}
        \vspace{-10pt}
        \caption{\textbf{Quantitative results on the MPII dataset} using the PCKh metric. We achieve state of the art performance and outperform significantly on difficult parts such as the ankle.}
        \label{fig:mpi_results}
\end{figure*}

\begin{figure*}[ht!]
    \centering
    \includegraphics[width=0.965\linewidth]{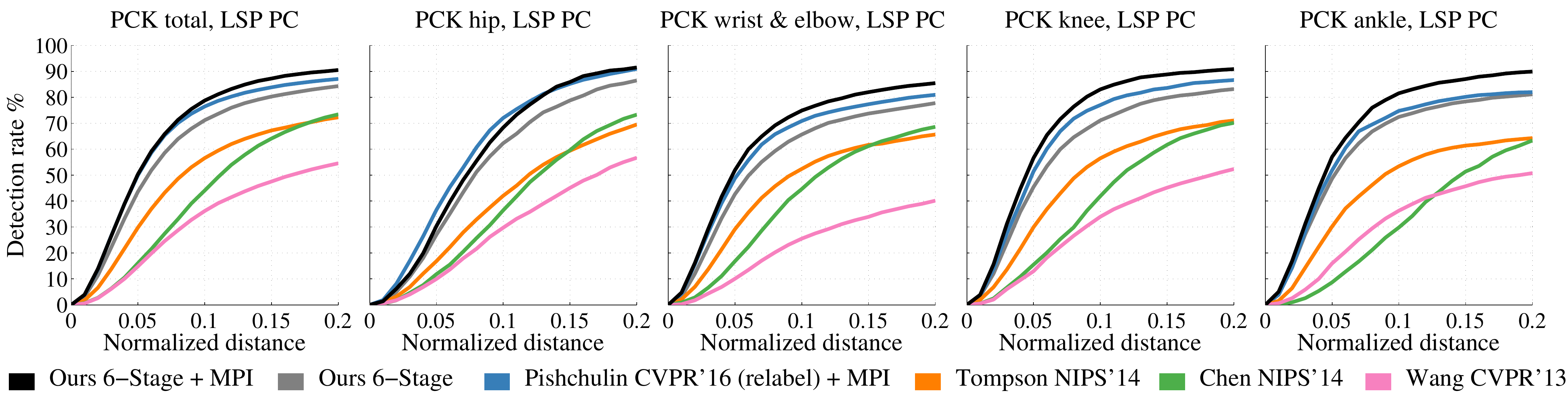}
    \vspace{-10pt}
    \caption{\textbf{Quantitative results on the LSP dataset} using the PCK metric. Our method again achieves state of the art performance and has a significant advantage on challenging parts.}
    \label{fig:leeds_oc_quant}
\end{figure*}

\minisection{Performance across stages.}
We show a comparison of performance across each stage on the LSP dataset (PC) in Figure \ref{fig:compare_stages}. We show that the performance increases monotonically until 5 stages, as the predictors in subsequent stages make use of contextual information in a large receptive field on the previous stage beliefs maps to resolve confusions between parts and background. We see diminishing returns at the 6th stage, which is the number we choose for reporting our best results in this paper for LSP and MPII datasets.


\subsection{Datasets and Quantitative Analysis}
\label{subset:Quantitative}

In this section we present our numerical results in various standard benchmarks including the MPII, LSP, and FLIC datasets. To have normalized input samples of $368 \times 368$ for training, we first resize the images to roughly make the samples into the same scale, and then
crop or pad the image according to the center positions and rough scale estimations provided in the datasets if available. In datasets such as LSP without these information, we estimate them according to joint positions or image sizes. For testing, we perform similar resizing and cropping (or padding), but estimate center position and scale only from image sizes when necessary. In addition, we merge the belief maps from different scales (perturbed around the given one) for final predictions, to handle the inaccuracy of the given scale estimation.

We define and implement our model using the \emph{Caffe} \cite{jia2014caffe} libraries for deep learning. We publicly release the source code and details on the architecture, learning parameters, design decisions and data augmentation to ensure full reproducibility.\footnote{\url{https://github.com/CMU-Perceptual-Computing-Lab/convolutional-pose-machines-release}}

\begin{figure*}[ht!]
    \centering
    \includegraphics[width=0.98\linewidth]{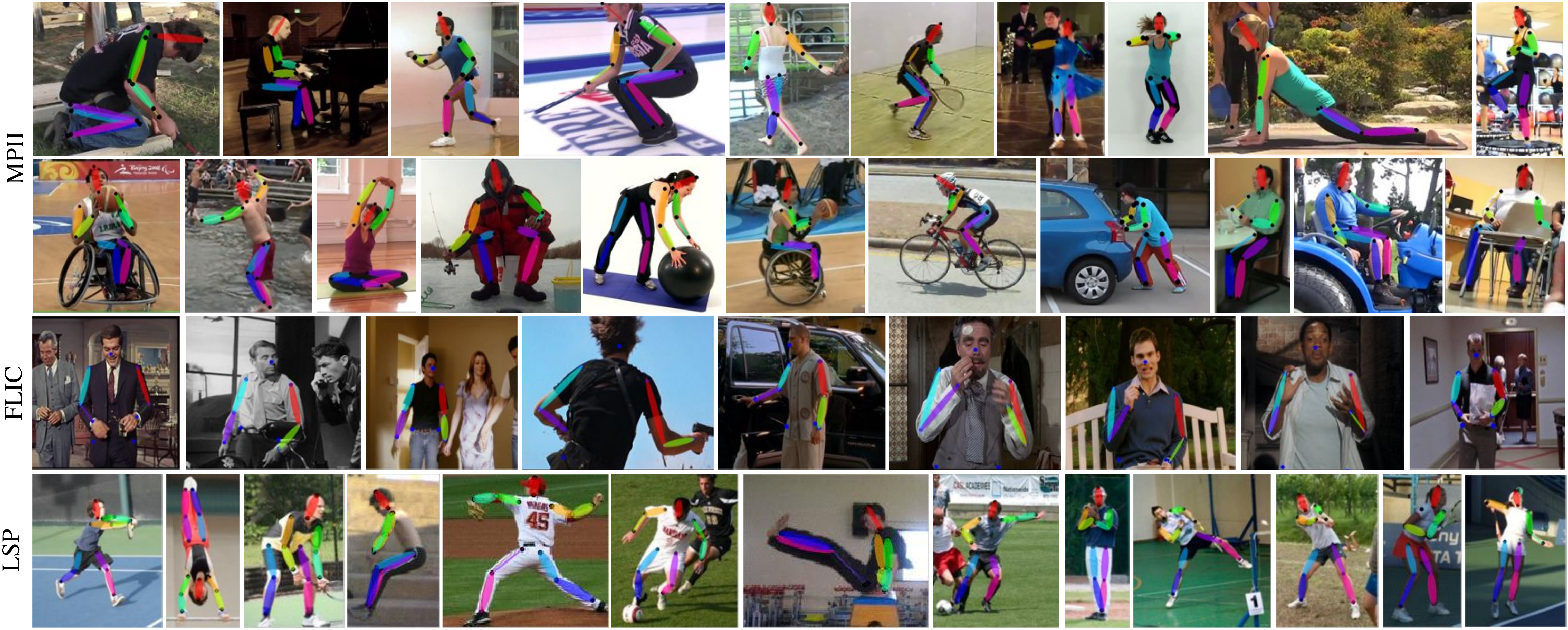}
    \vspace{-5pt}
    \caption{\textbf{Qualitative results} of our method on the MPII, LSP and FLIC datasets respectively.  We see that the method is able to handle non-standard poses and resolve ambiguities between symmetric parts for a variety of different relative camera views.}
    \label{fig:leeds_qualitative}
\end{figure*}

\begin{figure}[ht!]
    \centering
    \includegraphics[width=0.965\linewidth]{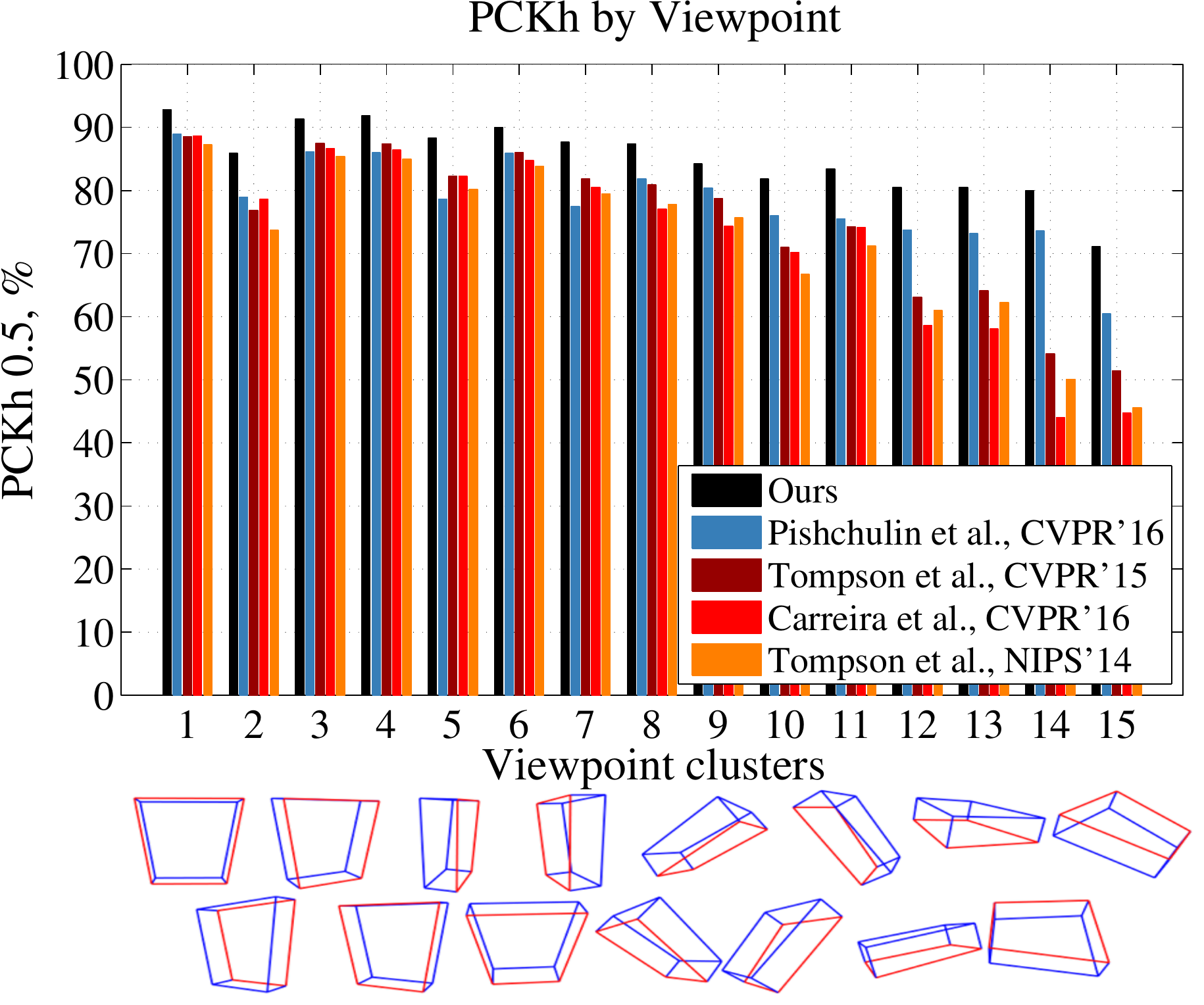}
    \vspace{-10pt}
    \caption{\textbf{Comparing PCKh-$0.5$ across various viewpoints in the MPII dataset.} Our method is significantly better in all the viewpoints.}
    \label{fig:mpi_vp}
\end{figure}

\minisection{MPII Human Pose Dataset}. We show in Figure \ref{fig:mpi_results} our results on the MPII Human Pose dataset \cite{andriluka14cvpr} which consists more than 28000 training samples. 
We choose to randomly augment the data with rotation degrees in $[-40^{\circ},40^{\circ}]$, scaling with factors in $[0.7,1.3]$, and horizonal flipping.
The evaluation is based on PCKh metric \cite{andriluka14cvpr} where the error tolerance is normalized with respect to head size of the target. Because there often are multiple people in the proximity of the interested person (rough center position is given in the dataset), we made two sets of ideal belief maps for training: 
one includes all the peaks for every person appearing in the proximity of the primary subject and the second type where we only place peaks for the primary subject.  We supply the first set of belief maps to the loss layers in the first stage as the initial stage only relies on local image evidence to make predictions. We supply the second type of belief maps to the loss layers of all subsequent stages.
We also find that supplying to all subsequent stages an additional heat-map with a Gaussian peak indicating center of the primary subject is beneficial.

\vspace{-3pt}

Our total PCKh-$0.5$ score achieves state of the art at $87.95\%$ ($88.52\%$ when adding LSP training data), which is $6.11\%$ higher than the closest competitor, and it is noteworthy that on the ankle (the most challenging part), our PCKh-$0.5$ score is $78.28\%$ ($79.41\%$ when adding LSP training data), which is $10.76\%$ higher than the closest competitor. This result shows the capability of our model to capture long distance context given ankles are the farthest parts from head and other more recognizable parts. Figure~\ref{fig:mpi_vp} shows our accuracy is also consistently significantly higher than other methods across various view angles defined in~\cite{andriluka14cvpr}, especially in those challenging non-frontal views.
In summary, our method improves the accuracy in all parts, over all precisions, across all view angles, and is the first one achieving such high accuracy without any pre-training from other data, or post-inference parsing with hand-design priors or initialization of such a structured prediction task as in \cite{tompson2014joint,pishchulin2015deepcut}. Our methods also does not need another module dedicated to location refinement as in \cite{tompson2015cvpr} to achieve great high-precision accuracy with a stride-8 network.


\vspace{-3pt}
\minisection{Leeds Sports Pose (LSP) Dataset.} We evaluate our method on the Extended Leeds Sports Dataset \cite{Johnson11} that consists of 11000 images for training and 1000 images for testing.
We trained on person-centric (PC) annotations and evaluate our method using the Percentage Correct Keypoints (PCK) metric \cite{yang2013articulated}.
Using the same augmentation scheme as for the MPI dataset, our model again achieves state of the art at $84.32\%$ ($90.5\%$ when adding MPII training data). Note that adding MPII data here significantly boosts our performance, due to its labeling quality being much better than LSP. Because of the noisy label in the LSP dataset, Pishchulin et al. \cite{pishchulin2015deepcut} reproduced the dataset with original high resolution images and better labeling quality.

\begin{figure}[ht!]
    \centering
    \includegraphics[width=\columnwidth]{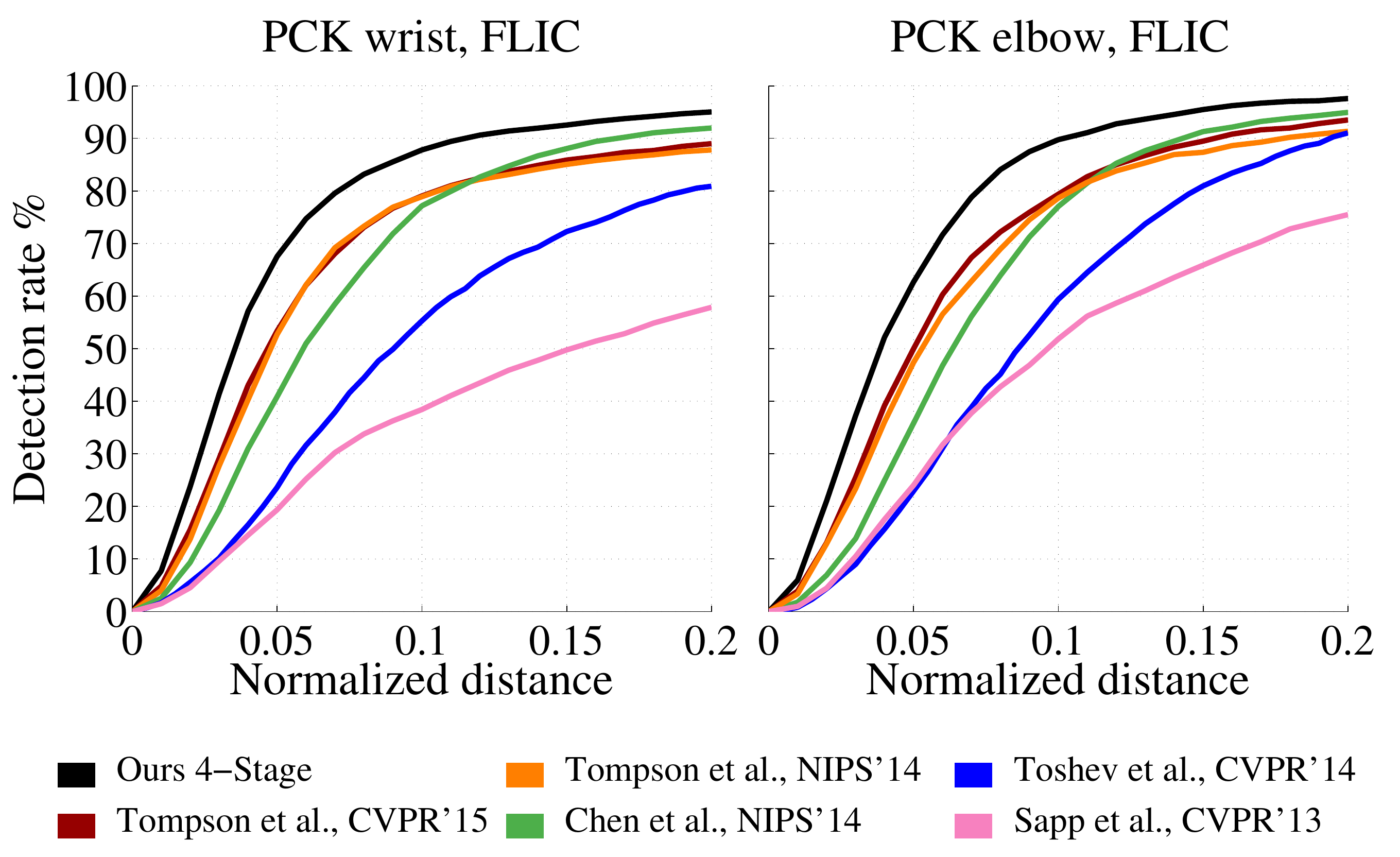}
    \vspace{-20pt}
    \caption{\textbf{Quantitative results on the FLIC dataset} for the elbow and wrist joints with a 4-stage  CPM. We outperform all competing methods.}
    \label{fig:flic_quant}
\end{figure}

\minisection{FLIC Dataset}. We evaluate our method on the FLIC Dataset \cite{sappmodec} which consists of 3987 images for training and 1016 images for testing. 
We report accuracy as per the metric introduced in Sapp et al. \cite{sappmodec} for the elbow and wrist joints in Figure \ref{fig:flic_quant}. Again, we outperform all prior art at PCK@0.2 with $97.59\%$ on elbows and $95.03\%$ on wrists. In higher precision region our advantage is even more significant: $14.8$ percentage points on wrists and $12.7$ percentage points on elbows at PCK@0.05, and $8.9$ percentage points on wrists and $9.3$ percentage points on elbows at PCK@0.1.

\vspace{-10pt}
\section{Discussion}
\vspace{-5pt}
Convolutional pose machines provide an end-to-end architecture for tackling structured prediction problems in computer vision without the need for graphical-model style inference. We showed that a sequential architecture composed of convolutional networks is capable of implicitly learning a spatial models for pose by communicating increasingly refined uncertainty-preserving beliefs between stages. Problems with spatial dependencies between variables arise in multiple domains of computer vision such as semantic image labeling, single image depth prediction and object detection and future work will involve extending our architecture to these problems.
%
Our approach achieves state of the art accuracy on all primary benchmarks, however we do observe failure cases mainly when multiple people are in close proximity. Handling multiple people in a single end-to-end architecture is also a challenging problem and an interesting avenue for future work.
%

{\small
\bibliographystyle{ieee}
\bibliography{posebib}

\begin{thebibliography}{10}\itemsep=-1pt

\bibitem{andriluka14cvpr}
M.~Andriluka, L.~Pishchulin, P.~Gehler, and B.~Schiele.
\newblock 2{D} human pose estimation: New benchmark and state of the art
  analysis.
\newblock In {\em CVPR}, 2014.

\bibitem{Andriluka2009}
M.~Andriluka, S.~Roth, and B.~Schiele.
\newblock Pictorial structures revisited: People detection and articulated pose
  estimation.
\newblock In {\em CVPR}, 2009.

\bibitem{Andriluka2010}
M.~Andriluka, S.~Roth, and B.~Schiele.
\newblock Monocular 3{D} pose estimation and tracking by detection.
\newblock In {\em CVPR}, 2010.

\bibitem{bengio1994learning}
Y.~Bengio, P.~Simard, and P.~Frasconi.
\newblock Learning long-term dependencies with gradient descent is difficult.
\newblock {\em IEEE Transactions on Neural Networks}, 1994.

\bibitem{bradley2010learning}
D.~Bradley.
\newblock {\em Learning In Modular Systems}.
\newblock PhD thesis, Robotics Institute, Carnegie Mellon University,
  Pittsburgh, PA, 2010.

\bibitem{carreia2015human}
J.~Carreira, P.~Agrawal, K.~Fragkiadaki, and J.~Malik.
\newblock Human pose estimation with iterative error feedback.
\newblock {\em arXiv preprint arXiv:1507.06550}, 2015.

\bibitem{Chen_NIPS14}
X.~Chen and A.~Yuille.
\newblock Articulated pose estimation by a graphical model with image dependent
  pairwise relations.
\newblock In {\em NIPS}, 2014.

\bibitem{Dantone2013}
M.~Dantone, J.~Gall, C.~Leistner, and L.~Van~Gool.
\newblock Human pose estimation using body parts dependent joint regressors.
\newblock In {\em CVPR}, 2013.

\bibitem{fh2005pictorial}
P.~Felzenszwalb and D.~Huttenlocher.
\newblock Pictorial structures for object recognition.
\newblock In {\em IJCV}, 2005.

\bibitem{glorot2010understanding}
X.~Glorot and Y.~Bengio.
\newblock Understanding the difficulty of training deep feedforward neural
  networks.
\newblock In {\em AISTATS}, 2010.

\bibitem{he15arxiv}
K.~He, X.~Zhang, S.~Ren, and J.~Sun.
\newblock Deep residual learning for image recognition.
\newblock {\em arXiv preprint arXiv:1512.03385}, 2015.

\bibitem{hochreiter2001gradient}
S.~Hochreiter, Y.~Bengio, P.~Frasconi, and J.~Schmidhuber.
\newblock Gradient flow in recurrent nets: the difficulty of learning long-term
  dependencies.
\newblock {\em A Field Guide to Dynamical Recurrent Neural Networks, IEEE
  Press}, 2001.

\bibitem{jia2014caffe}
Y.~Jia, E.~Shelhamer, J.~Donahue, S.~Karayev, J.~Long, R.~Girshick,
  S.~Guadarrama, and T.~Darrell.
\newblock Caffe: Convolutional architecture for fast feature embedding.
\newblock {\em arXiv preprint arXiv:1408.5093}, 2014.

\bibitem{johnson2010clustered}
S.~Johnson and M.~Everingham.
\newblock Clustered pose and nonlinear appearance models for human pose
  estimation.
\newblock In {\em BMVC}, 2010.

\bibitem{Johnson11}
S.~Johnson and M.~Everingham.
\newblock Learning effective human pose estimation from inaccurate annotation.
\newblock In {\em CVPR}, 2011.

\bibitem{karlinsky2012using}
L.~Karlinsky and S.~Ullman.
\newblock Using linking features in learning non-parametric part models.
\newblock In {\em ECCV}, 2012.

\bibitem{kiefel2014human}
M.~Kiefel and P.~V. Gehler.
\newblock Human pose estimation with fields of parts.
\newblock In {\em ECCV}. 2014.

\bibitem{krizhevsky2012imagenet}
A.~Krizhevsky, I.~Sutskever, and G.~E. Hinton.
\newblock Imagenet classification with deep convolutional neural networks.
\newblock In {\em NIPS}, 2012.

\bibitem{lan2005beyond}
X.~Lan and D.~Huttenlocher.
\newblock Beyond trees: Common-factor models for 2{D} human pose recovery.
\newblock In {\em ICCV}, 2005.

\bibitem{lee2014deeply}
C.-Y. Lee, S.~Xie, P.~Gallagher, Z.~Zhang, and Z.~Tu.
\newblock Deeply-supervised nets.
\newblock In {\em AISTATS}, 2015.

\bibitem{long_shelhamer_fcn}
J.~Long, E.~Shelhamer, and T.~Darrell.
\newblock Fully convolutional networks for semantic segmentation.
\newblock In {\em CVPR}, 2015.

\bibitem{munoz2010}
D.~Munoz, J.~Bagnell, and M.~Hebert.
\newblock Stacked hierarchical labeling.
\newblock In {\em ECCV}, 2010.

\bibitem{ouyang2014multi}
W.~Ouyang, X.~Chu, and X.~Wang.
\newblock Multi-source deep learning for human pose estimation.
\newblock In {\em CVPR}, 2014.

\bibitem{pfister2015flowing}
T.~Pfister, J.~Charles, and A.~Zisserman.
\newblock Flowing convnets for human pose estimation in videos.
\newblock In {\em ICCV}, 2015.

\bibitem{Pinheiro14recurrentconvolutional}
P.~Pinheiro and R.~Collobert.
\newblock Recurrent convolutional neural networks for scene labeling.
\newblock In {\em ICML}, 2014.

\bibitem{pishchulin2013poselet}
L.~Pishchulin, M.~Andriluka, P.~Gehler, and B.~Schiele.
\newblock Poselet conditioned pictorial structures.
\newblock In {\em CVPR}, 2013.

\bibitem{pishchulin13iccv}
L.~Pishchulin, M.~Andriluka, P.~Gehler, and B.~Schiele.
\newblock Strong appearance and expressive spatial models for human pose
  estimation.
\newblock In {\em ICCV}, 2013.

\bibitem{pishchulin2015deepcut}
L.~Pishchulin, E.~Insafutdinov, S.~Tang, B.~Andres, M.~Andriluka, P.~Gehler,
  and B.~Schiele.
\newblock Deepcut: Joint subset partition and labeling for multi person pose
  estimation.
\newblock {\em arXiv preprint arXiv:1511.06645}, 2015.

\bibitem{Ramakrishna2014posemachines}
V.~Ramakrishna, D.~Munoz, M.~Hebert, J.~Bagnell, and Y.~Sheikh.
\newblock {P}ose {M}achines: {A}rticulated {P}ose {E}stimation via {I}nference
  {M}achines.
\newblock In {\em ECCV}, 2014.

\bibitem{ramanan2005strike}
D.~Ramanan, D.~A. Forsyth, and A.~Zisserman.
\newblock Strike a {P}ose: {T}racking people by finding stylized poses.
\newblock In {\em CVPR}, 2005.

\bibitem{ross2011}
S.~Ross, D.~Munoz, M.~Hebert, and J.~Bagnell.
\newblock Learning message-passing inference machines for structured
  prediction.
\newblock In {\em CVPR}, 2011.

\bibitem{sappmodec}
B.~Sapp and B.~Taskar.
\newblock {MODEC}: {M}ultimodal {D}ecomposable {M}odels for {H}uman {P}ose
  {E}stimation.
\newblock In {\em CVPR}, 2013.

\bibitem{sigal2006measure}
L.~Sigal and M.~Black.
\newblock Measure locally, reason globally: Occlusion-sensitive articulated
  pose estimation.
\newblock In {\em CVPR}, 2006.

\bibitem{steward2015endtoend}
R.~Stewart and M.~Andriluka.
\newblock End-to-end people detection in crowded scenes.
\newblock {\em arXiv preprint arXiv:1506.04878}, 2015.

\bibitem{sun2011articulated}
M.~Sun and S.~Savarese.
\newblock Articulated part-based model for joint object detection and pose
  estimation.
\newblock In {\em ICCV}, 2011.

\bibitem{szegedy2014going}
C.~Szegedy, W.~Liu, Y.~Jia, P.~Sermanet, S.~Reed, D.~Anguelov, D.~Erhan,
  V.~Vanhoucke, and A.~Rabinovich.
\newblock Going deeper with convolutions.
\newblock {\em arXiv preprint arXiv:1409.4842}, 2014.

\bibitem{tian2012exploring}
Y.~Tian, C.~L. Zitnick, and S.~G. Narasimhan.
\newblock Exploring the spatial hierarchy of mixture models for human pose
  estimation.
\newblock In {\em ECCV}. 2012.

\bibitem{tompson2015cvpr}
J.~Tompson, R.~Goroshin, A.~Jain, Y.~LeCun, and C.~Bregler.
\newblock Efficient object localization using convolutional networks.
\newblock In {\em CVPR}, 2015.

\bibitem{tompson2014joint}
J.~Tompson, A.~Jain, Y.~LeCun, and C.~Bregler.
\newblock Joint training of a convolutional network and a graphical model for
  human pose estimation.
\newblock In {\em NIPS}, 2014.

\bibitem{toshev2013deeppose}
A.~Toshev and C.~Szegedy.
\newblock Deep{P}ose: Human pose estimation via deep neural networks.
\newblock In {\em CVPR}, 2013.

\bibitem{tu2010PAMI}
Z.~Tu and X.~Bai.
\newblock Auto-context and its application to high-level vision tasks and 3d
  brain image segmentation.
\newblock In {\em TPAMI}, 2010.

\bibitem{wang2008multiple}
Y.~Wang and G.~Mori.
\newblock Multiple tree models for occlusion and spatial constraints in human
  pose estimation.
\newblock In {\em ECCV}, 2008.

\bibitem{yang2011articulated}
Y.~Yang and D.~Ramanan.
\newblock Articulated pose estimation with flexible mixtures-of-parts.
\newblock In {\em CVPR}, 2011.

\bibitem{yang2013articulated}
Y.~Yang and D.~Ramanan.
\newblock Articulated human detection with flexible mixtures of parts.
\newblock In {\em TPAMI}, 2013.

\end{thebibliography}
}

\end{document}